\documentclass{article}

\usepackage{graphicx}%
\usepackage{multirow}%
\usepackage{amsmath,amssymb,amsfonts}%
\usepackage{amsthm}%
\usepackage{mathrsfs}%
\usepackage[title]{appendix}%
\usepackage{xcolor}%
\usepackage{textcomp}%
\usepackage{manyfoot}%
\usepackage{booktabs, comment}%
\usepackage{listings}%

\usepackage{subcaption}
\usepackage{longtable}
\usepackage{tabularx,booktabs}
\usepackage{multicol, multirow}
\usepackage{amssymb,pifont} 
\usepackage{adjustbox} 

\providecommand{\keywords}[1]
{
  \small	
  \textbf{Keywords---} #1
}

\usepackage[linesnumbered,ruled,vlined]{algorithm2e}
\SetKwInput{TrainingPhase}{Training Phase}
\SetKwInput{TestPhase}{Test Phase}
\SetKwInput{Given}{Given}
\SetKwInput{Hyperparam}{Hyper-parameters}
\SetKwInput{Procedure}{Procedure}
\SetKwInput{If}{if}
\SetKwInput{Else}{else}
\SetKwInput{Return}{Return}

\title{Heterogeneous Random Forest}

\author{
    Ye-eun Kim \textsuperscript{1} \hspace{1.5cm} 
    Seoung Yun Kim \textsuperscript{2} \hspace{1.5cm} 
    Hyunjoong Kim\textsuperscript{1}
}

\begin{document}

\maketitle
\footnotetext[1]{\textsuperscript{}Department of Statistics and Data Science, Yonsei University, Seoul, South Korea.}
\footnotetext[2]{\textsuperscript{}Business Data Division, GS Retail, Seoul, South Korea.}
\begin{abstract}
Random forest (RF) stands out as a highly favored machine learning approach for classification problems. The effectiveness of RF hinges on two key factors: the accuracy of individual trees and the diversity among them. In this study, we introduce a novel approach called heterogeneous RF (HRF), designed to enhance tree diversity in a meaningful way. This diversification is achieved by deliberately introducing heterogeneity during the tree construction. Specifically, features used for splitting near the root node of previous trees are assigned lower weights when constructing the feature sub-space of the subsequent trees. As a result, dominant features in the prior trees are less likely to be employed in the next iteration, leading to a more diverse set of splitting features at the nodes. 
Through simulation studies, it was confirmed that the HRF method effectively mitigates the selection bias of trees within the ensemble, increases the diversity of the ensemble, and demonstrates superior performance on datasets with fewer noise features.
To assess the comparative performance of HRF against other widely adopted ensemble methods, we conducted tests on 52 datasets, comprising both real-world and synthetic data. HRF consistently outperformed other ensemble methods in terms of accuracy across the majority of datasets.
\end{abstract}

\section{Introduction}
Random forest (RF, \cite{Breiman2001}) is one of the most prevalent machine learning techniques for classification and regression problems. It is a tree-based ensemble method that aggregates the results of numerous decision trees to generate prediction outcomes. RF employs a strategy to induce diversity among classifiers. This is accomplished by choosing a random subset of candidate features from the pool of available features and then splitting a node of the tree based on the most optimal variable within this subset.

The performance of ensemble methods is influenced by two primary factors: the individual strength of the base learners and the correlation between them. In other words, if the trees themselves are effective classifiers, the forest will also demonstrate superior performance. Additionally, if the trees are more heterogeneous and diverse, the forest is likely to exhibit enhanced performance. Consequently, numerous studies have explored modifications to RF, with the objective of improving accuracy, diversity, or a combination of both.

Several studies have enhanced the performance of ensemble models by developing more sophisticated trees. Specifically, some approaches have focused on selecting only the most relevant variables or eliminating redundant ones. This can be seen as a feature selection technique aimed at improving the accuracy of individual trees(\cite{NANNI200911257}, \cite{KUMAR2023101959}, \cite{10.1007/978-3-642-41822-8_35}, \cite{Darst2018}, \cite{fuzzyForests}, \cite{ROKACH20081676}). In contrast, other research has explored the construction of oblique trees to simultaneously boost accuracy and diversity. These methods involve transforming features, such as through linear combinations, to create splits that are not aligned with the original feature axes\cite{1677518}. For instance, \cite{Chen2014} and \cite{rainforth2017canonical} proposed rotated trees based on techniques like LDA (Linear Discriminant Analysis), CLDA (Canonical LDA), and CCA (Canonical Correlation Analysis). However, these rotation-based methods often sacrifice the interpretability that is one of the key strengths of tree-based models.

Instead of focusing solely on the performance of individual trees, some approaches emphasize increasing the diversity among trees. A common strategy involves introducing additional randomness to reduce the correlation between trees. For instance, \cite{Han2020} demonstrated improved classification performance by enhancing diversity through bootstrapping at the node level. Similarly, \cite{Geurts2006} employed a method that selects candidate split points randomly, further promoting diversity across the ensemble.

This paper proposes a novel ensemble method called heterogeneous random forest (HRF). Tree diversity was induced by regulating the variables employed for tree splitting during the training phase. To accomplish this, we employed a weighted sampling approach to choose candidate features. Notably, weighted sampling has also been applied in other previous works, such as those by \cite{Xu2012}, \cite{Amar2008}, \cite{Ye2013} and \cite{Wang2018}.

The remainder of the article is organized as follows: Section~\ref{subsec:randomforests} provides an overview of random forest (RF). Sections~\ref{subsec:basic_idea} and \ref{subsec:feature_depth} provide an introduction to the fundamental concepts of HRF. 
The HRF algorithm is outlined in Section~\ref{subsec:HRF} with the aid of a simplified illustrative example. 
In Section~\ref{sec:Diversity of Trees}, a measure for assessing the diversity of trees in the ensemble was devised and introduced.
In Section~\ref{sec:Simulation}, we conduct an experiment on a simulation dataset comprising various input features to gain insight into the characteristics of HRF. Section~\ref{sec:empirical_evaluation} presents the empirical results, including the accuracy assessment. The results are then compared with those obtained from other ensemble methods. The following ensemble methods were considered: bagging \cite{Breiman1996}, random forest \cite{Breiman2001}, extremely randomized trees \cite{Geurts2006}, gradient boosting \cite{GB}, XGBoost \cite{xgb}, and CatBoost \cite{catboost}. 
Section~\ref{sec:conlusion} concludes this paper.

\section{Method}\label{sec:HRF}
\subsection{Random forest}
\label{subsec:randomforests}

Random forest (RF) is a type of bagging method that employs bootstrap sampling for each tree. The distinction between bagging and RF lies in the random selection of $m$ candidate features for splitting. At each node of the tree, a new random feature subset is constructed, and the optimal split is selected based on the goodness of split within the subset. RF employs this strategy to induce diversity among individual trees, introducing additional randomness. The detailed algorithm is described in Algorithm~\ref{alg:RF}. In general, the most commonly used value for hyper-parameter $m$ are $\lfloor\sqrt{p}\rfloor$ and $\lfloor p/3 \rfloor$ for classification and regression, respectively.

\begin{algorithm}[H]
\setcounter{AlgoLine}{0}
\caption{Random Forest Algorithm}
\label{alg:RF}
\KwIn{
\begin{itemize}
    \item $D$ : training set with $n$ instances, $p$ features and a target variable
\end{itemize}
}
\Hyperparam{
\begin{itemize}
    \item $B$ : the number of trees in an ensemble
    \item $m \in \{1, \ldots, p\}$ : the number of candidate features to be selected at each node
\end{itemize}}
\KwOut{}
\begin{itemize}
    \item $H=(h_1,...,h_B)$ : the trained forest
\end{itemize}

\medskip

\TrainingPhase{\\
\For{$b = 1$ \textbf{to} $B$}{
    Generate bootstrapped sample $D^*_b$ from training set $D$
    
    Grow a tree using bootstrapped sample $D^*_b$
    
    \While{all stopping rules are not met}{
        Randomly select $m$ candidate features
        
        Find the best split features and cut-point using $m$ features
        
        Send down the data using the selected features and cut-point
    }
}
}

\medskip

\TestPhase{Aggregate the $B$ trained classifiers using simple majority vote. The predicted result for a new sample $x$ from model $h$ is given as:
    \[
    h(x) \gets \arg\max_k \sum_{b=1}^{B}I\left[h_b(x) = k\right], k \in \{1,...,K\},
    \]
where $K$ denotes the number of classes in the target variable.
}
\end{algorithm}

\subsection{Basic idea of new method}
\label{subsec:basic_idea}

The new method employs a memory system that retains the structural characteristics of previous trees. This stored information is then leveraged to generate more diverse future trees.

Because of the hierarchical nature of decision trees, splits at shallow depths (near root node) have broad implications for the entire space, while splits at deeper depths exert more localized effects and have a less impact on the overall tree structure~\cite{201700141}.

Therefore, to effectively enhance diversity, it is more crucial to manage the features selected for splitting near the root node than those at the lower levels.
We intend to ensure that features picked close to the root node in earlier trees are less frequently selected in subsequent trees.
This can be achieved through weighted sampling, incorporating information from previous trees when selecting the feature subset for the next tree. 
This procedure differs from RF, which apply simple random sampling without considering the results of previous trees.

\subsection{Feature depth}
\label{subsec:feature_depth}

\begin{figure}
    \centering
    \begin{subfigure}{0.33\textwidth}
        \centering
        \includegraphics[width=\textwidth]{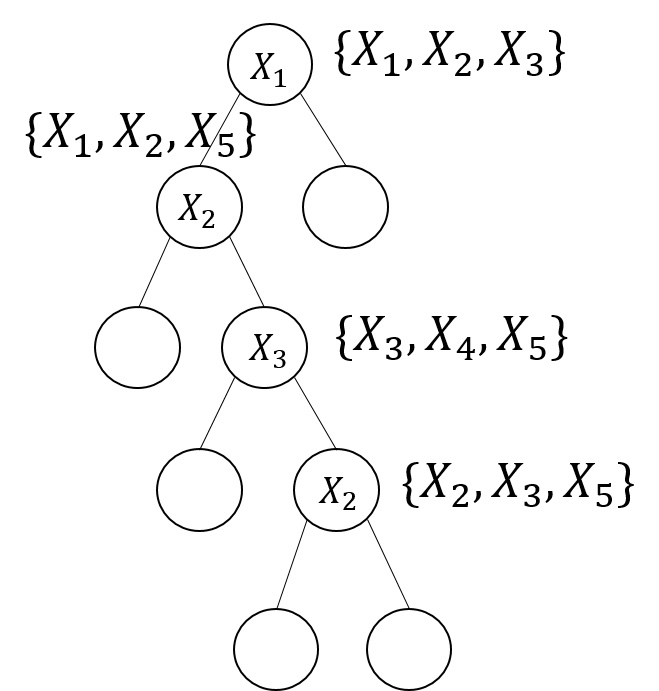}
        \caption{Tree 1: $h_1(x)$}    
    \end{subfigure}
    \begin{subfigure}{0.33\textwidth}
        \centering
        \includegraphics[width=\textwidth]{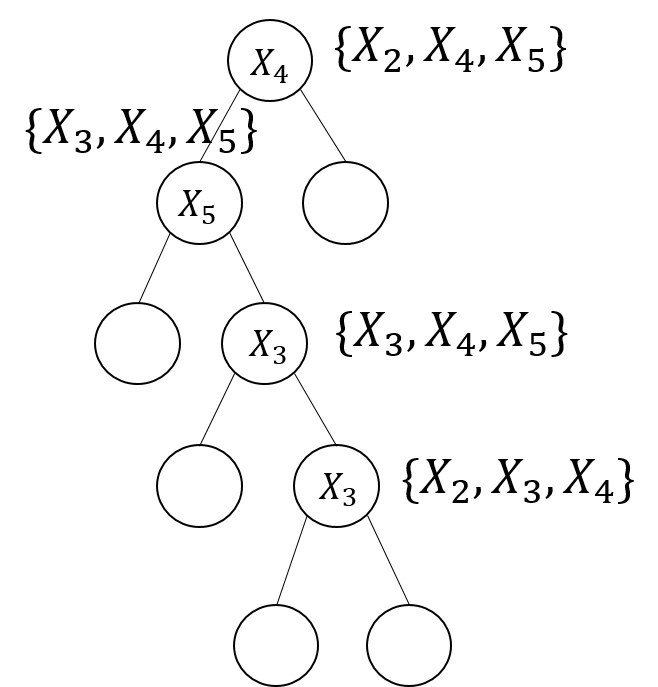}
        \caption{Tree 2: $h_2(x)$}    
    \end{subfigure}
    \caption{Decision tree examples. The feature inside the node is the split variable and the features in brackets next to the node represent candidate features}
    \label{trees_example}
\end{figure}

Let us consider an individual tree within an ensemble. We define a feature depth as the earliest level a feature is used. Formally, the feature depth $d_{bj}^{\beta}$ is defined as 
\begin{equation}
\label{equ:feature_depth}
    d_{b,j}^{\beta} = \begin{cases}
\begin{aligned}
    & \text{minimum depth that $X_j$ used for split,} && \text{if }X_j \text{ is used in $b$-th tree} \\
    & M_b -1+\beta, && \text{otherwise,} 
\end{aligned}
\end{cases}
\end{equation}
where $j=1,2,...,p$, $b=1,2,...,B$, $M_b$ represents the depth of $h_b$, and $\beta$ is a hyper-parameter for unused features typically set to 1. 

To illustrate, consider the case where the first decision tree, denoted by $h_1(x)$, was built following the structure shown in Figure~\ref{trees_example}(a). The $X_1$ feature is selected as the splitting criterion at the root node, thus it is assigned a value of 0 for the feature depth. In the case of features that appear more than once, the smallest number is recorded, that is, the number closest to the root node. It can be observed that the  $X_2$ feature is present in both the first and the third levels of $h_1(x)$. Consequently, it is assigned a value of $1$. 
In this example, $M_b$ is equal to $4$. In the event that both $X_4$ and $X_5$ are not selected, they will each have a value of $4$ if the parameter $\beta$ is set to $1$. The same process can be applied to calculate the feature depth of the second tree.


\subsection{Heterogeneous random forest}
\label{subsec:HRF}

A shallower feature depth indicates a more significant feature in the previous tree. Consequently, feature depth can be employed as a selection weight when identifying potential feature sets. This implies that features that were used as influential factors in the preceding tree will be employed with diminished frequency in the subsequent trees. 

\begin{table}
    \begin{subtable}{0.45\textwidth}
        \centering
         \begin{tabular}{cccccc}
            \toprule
             $\boldsymbol{d}$ & $X_1$ & $X_2$ & $X_3$ & $X_4$ & $X_5$ \\
             \midrule
             $\boldsymbol{d}_1$ & 0 & 1 & 2 & 4 & 4 \\
             $\boldsymbol{d}_2$ & 4 & 4 & 2 & 0 & 1 \\
             \bottomrule
        \end{tabular}
        \caption{feature depth of $h_1(x)$ and $h_2(x)$ when $\beta=1$}        
    \end{subtable}
    \hspace{1cm}
    \begin{subtable}{0.45\textwidth}
    \centering
     \begin{tabular}{cccccc}
        \toprule
         $\boldsymbol{D}$ & $X_1$ & $X_2$ & $X_3$ & $X_4$ & $X_5$ \\
         \midrule
         $\boldsymbol{D}_1$ & 0 & 1 & 2 & 4 & 4 \\
         $\boldsymbol{D}_2$ & 4 & 4.5 & 3 & 2 & 3 \\
         \bottomrule
    \end{tabular}
    \caption{Cumulative feature depth when $\alpha=0.5$}  \end{subtable}
    \centering
    \begin{subtable}{0.6\textwidth}
     \begin{tabular}{cccccc}
        \toprule
         $\boldsymbol{w}$ & $X_1$ & $X_2$ & $X_3$ & $X_4$ & $X_5$ \\
         \midrule
         $\boldsymbol{w}_2$ & 0/11 & 1/11 & 2/11 & 4/11 & 4/11 \\
         $\boldsymbol{w}_3$ & 4/16.5 & 4.5/16.5 & 3/16.5 & 2/16.5 & 3/16.5 \\
         \bottomrule
    \end{tabular}
    \caption{feature weights: $\boldsymbol{w}_2$ for $h_2(x)$ and $\boldsymbol{w}_3$ for $h_3(x)$}
    \end{subtable}
    \caption{The procedure to update feature weight}
    \label{tab:procedure_feature_weight}
\end{table}

The example illustrated in Figure~\ref{trees_example} will continue to be used for explanatory purposes. 
In Table~\ref{tab:procedure_feature_weight}(a), the feature depth of the first tree, $\boldsymbol{d}_1$, is recorded. The next step is to calculate the cumulative depth $\boldsymbol{D}_1$ for each variable from all of the previous trees, but since this tree is the first one, the cumulative depths as given in Table~\ref{tab:procedure_feature_weight}(b) are just equal to the tree depths $\boldsymbol{d}_1$. The feature weights for the second tree, $\boldsymbol{w}_2$, are determined by dividing the cumulative depth of each feature by the total sum of all cumulative depths of the first tree as in Table~\ref{tab:procedure_feature_weight}(c).

Moving on to constructing the second tree in Figure~\ref{trees_example}(b), the feature subset for each node of the second tree is sampled by using the weights $\boldsymbol{w}_2$ in Table~\ref{tab:procedure_feature_weight}(c). Note that the feature $X_1$, which was used at the root node of the first tree, does not appear in the second tree at all. In contrast, $X_4$ and $X_5$, which were not selected in the first tree, both appear near the root node of the second tree.

The next step is to recalculate and update the weights for each variable. As previously indicated, the depths for each variable in the second tree should be recorded (denoted as $\boldsymbol{d}_2$), and the cumulative depths (denoted as $\boldsymbol{D}_2$) should then be calculated. However, we intend to use a more refined method rather than simply adding the depths of each tree to obtain the cumulative depth. The depth of the preceding tree is multiplied by the hyper-parameter $\alpha$ and then added to the current tree depth. For example, when $\alpha=0.5$, $X_2$ receives a cumulative depth of $4.5
$, calculated as $1 \times 0.5 + 4$.

The hyper-parameter $\alpha$ can be interpreted as the memory parameter. It controls the influence of the previous trees, or how much the next tree remembers its predecessors. If $\alpha$ is equal to $1$, then all information is fully retained. The $b$-th tree is influenced to the same extent by the first tree as it is by the immediate predecessor. As the value of $\alpha$ approaches $0$, the influence of the preceding trees is rapidly diminished. Only the more recent trees will exert any influence on future ones. The cumulative depth resulting from the $b$-th construction can be expressed as follows:
\begin{align}
\label{equ:cum_depth}
\boldsymbol{D}_b &= \boldsymbol{d}_b + \alpha\boldsymbol{d}_{b-1} + \alpha^2\boldsymbol{d}_{b-2} + \cdots + \alpha^{b-1}\boldsymbol{d}_1\\
& =  \boldsymbol{d}_b +\sum_{k=1}^{b-1}\alpha^k\boldsymbol{d}_{b-k},
\end{align}
where $\boldsymbol{d}_{b} = [d_{b,1}^{\beta},\cdots,d_{b,p}^{\beta}]$ is a vector of feature depths.

Following the calculation of $\boldsymbol{D}_b$, the updated weights are obtained by Algorithm~\ref{alg:upd}, which are then used for the feature sampling weights of the subsequent tree. The hyper-parameter $\beta$ controls the degree of advantage that non-used features will have. If the value of $\beta$ is large, features that were not selected in previous trees will have higher weights in subsequent trees. 

Finally, the proposed heterogeneous random forest (HRF) algorithm is summarized in Algorithm~\ref{alg:hRF}.

\begin{algorithm}[H]
\setcounter{AlgoLine}{0}
\caption{Weight Updating Algorithm}\label{alg:upd}
\KwIn{
\begin{itemize}
    \item $\boldsymbol{D}_{b-1} = [D_{b-1,1},\cdots,D_{b-1,p}]$ : Cumulative feature depths
    \item $\boldsymbol{d}_{b} = [d_{b,1}^{\beta},\cdots,d_{b,p}^{\beta}]$ : feature depths
\end{itemize}}
\Hyperparam{
\begin{itemize}
    \item $\alpha\in(0,1)$ : How much will previous trees be reflected in
\end{itemize}
}
\KwOut{
\begin{itemize}
    \item $\boldsymbol{w}_{b+1}$ : The updated weight
\end{itemize}
}

\medskip

\Procedure{}
\eIf{$b = 1$}{
    $\boldsymbol{D}_{b}\gets \boldsymbol{d}_{b}$
    }{
    $\boldsymbol{D}_{b} \gets \boldsymbol{d}_{b}+\alpha\cdot\boldsymbol{D}_{b-1} $ }
$\boldsymbol{w}_{b+1} \gets \frac{1}{\sum_{j=1}^{p}{D_{b,j}}}\cdot\boldsymbol{D}_{b}$

\end{algorithm}

\begin{algorithm}[H]
\setcounter{AlgoLine}{0}
\caption{Heterogeneous Random Forest Algorithm}\label{alg:hRF}
\KwIn{
\begin{itemize}
    \item $D$ : training set with $n$ instances, $p$ features and a target variable
\end{itemize}}

\Hyperparam{
\begin{itemize}
    \item $B$ : the number of trees in an ensemble
    \item $m \in \{1, \ldots, p\}$: the number of candidate features to be selected at each node
    \item $\alpha\in(0,1)$ : 
    The degree of influence of past trees on future trees
    \item $\beta\in \mathbb{N}$ : The degree of advantage given to the non-used features
\end{itemize}
}

\medskip

\TrainingPhase{}
$\boldsymbol{w}_1 \gets \left[\frac{1}{p},...,\frac{1}{p}\right]$

\For{$b = 1$ \textbf{to} $B$}{
    Generate bootstrapped sample $D^*_b$ from training set $D$
    
    Grow a tree using bootstrapped sample $D^*_b$ 
    
    \While{all stopping rules are not met}{
        Randomly select $m$ candidate features with weight $\boldsymbol{w}_{b}$
        
        Find the best split features and cut-point using $m$ features
        
        Send down the data using the selected features and cut-point
        } 
        $\boldsymbol{d}_b\gets[d_{b,1}^{\beta},\cdots,d_{b,p}^{\beta}]$ from Equation (\ref{equ:feature_depth})
        
        $\boldsymbol{D}_b\gets[\alpha, \boldsymbol{d}_{b},\cdots,\boldsymbol{d}_{1}]$ from Equation (\ref{equ:cum_depth})
        
        $\boldsymbol{w}_{b+1} \gets \text{Weight update}$ by Algorithm \ref{alg:upd}

} 

\medskip

\TestPhase{Aggregate the $B$ trained classifiers using simple majority vote. The predicted for a new sample $x$ from model $h$ is given as: 
$$
    h(x) \gets \arg\max_k \sum_{b=1}^{B}I\left[h_b(x) = k\right], k \in \{1,...,K\},
$$
where $K$ denotes the number of classes in the target variable.}
\end{algorithm}

\subsection{New measure of diversity}\label{sec:Diversity of Trees}

Our proposed method, HRF, seeks to improve the diversity of trees within an ensemble. In this section, we introduce a new metric to measure tree dissimilarity, aimed at evaluating the diversity of HRF. Specifically, we devised a method to quantify the level of heterogeneity among the various tree structures generated in the ensemble.

\begin{figure}
    \centering
    \includegraphics[width=0.9\textwidth]{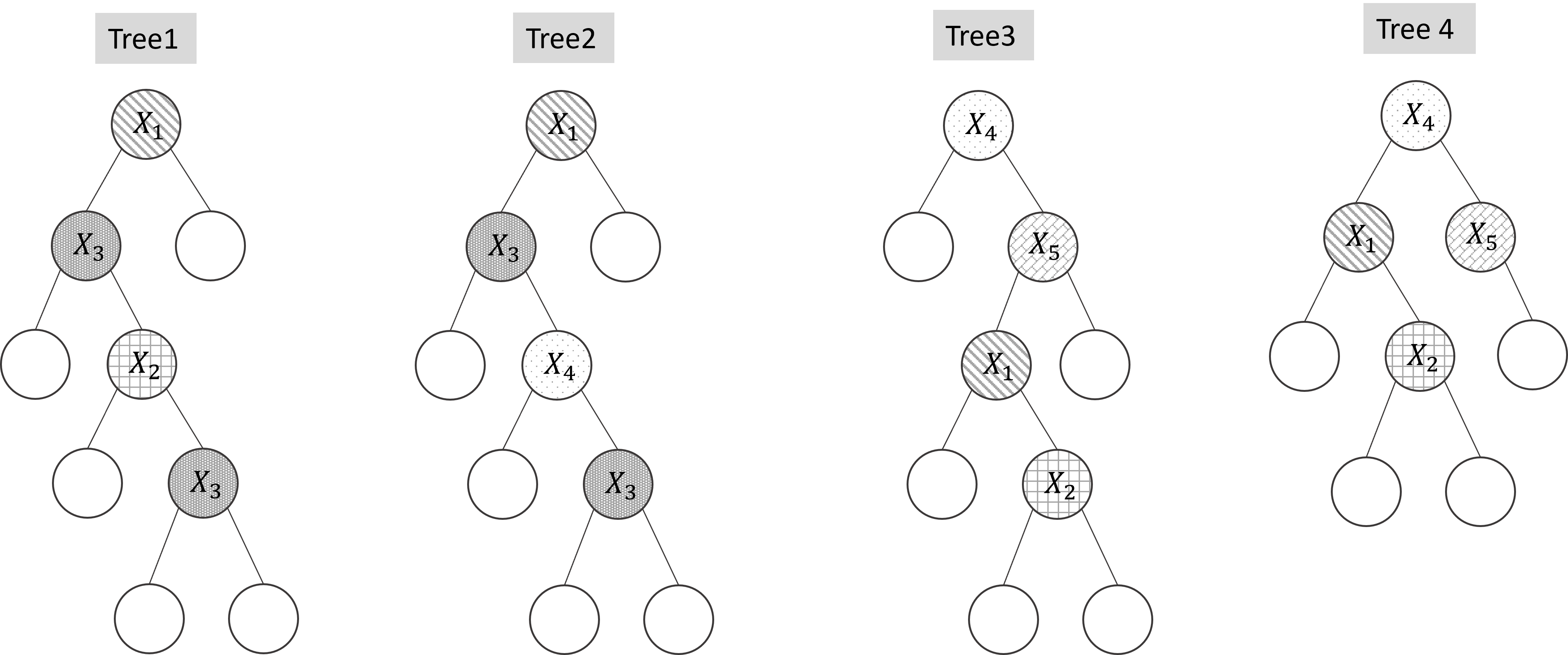}
    \caption{Trees for toy example}
    \label{fig:tree_samples4}
\end{figure}

The characteristics of a decision tree can be summarized by the feature depth, $\boldsymbol{d}_{b}$, which denotes the variable used for splitting at a given depth. We utilize this feature depth to define feature dominance, which can reflect the differences between two trees.

The dominance of a $j$-th feature in $b$-th classifier is defined as 
\begin{equation}
\label{equ:feature_dominance}
    I_{b,j}= \begin{cases}
\begin{aligned}
    & M_{b}^{*} + 1 - d_{b,j}^{\beta} && \text{if }X_j \text{ is used in $b$-th tree} \\
    & 0, && \text{otherwise,} 
\end{aligned}
\end{cases}
\end{equation}
 where $M^{*}_{b}$ denotes the maximum value of the feature depth of split variables. Unused variables have a value of 0, whereas variables that are employed in proximity to the root node are assigned a relatively high value. 
By concatenating the dominance of the $l$-th and $m$-th trees by row, $[\boldsymbol{I}_l , \boldsymbol{I}_m] $, a table consisting of 2 rows and $p$ columns is created. We can perform a $\chi^2$ test of homogeneity with this table.
A large $\chi^2$ test statistic means that the feature depths of each variable in the two trees are very different, and a small value means that they are similar.

Figure \ref{fig:tree_samples4} presents examples to illustrate the dissimilarity between trees. Trees 1 and 2 exhibit similarity, as do trees 3 and 4. 
Table \ref{tab:dissimiarity_toy_example}(a) presents the dominance of each feature, while Table \ref{tab:dissimiarity_toy_example}(b) displays the results of the $\chi^2$ test statistic, which measures the similarity between trees.

To facilitate comparison in cases where there are different degrees of freedom, one can transform the $\chi^2$ test statistic to a standard normal variable using the Wilson–Hilferty\cite{Hilferty} method. Ultimately, the dissimilarity between $h_l$ and $h_m$ is given by the following equation:
\begin{equation}
    DS(h_l,h_m) = \frac{\sqrt[3]{X/k} - (1-2/(9k))}{\sqrt{2/(9k)}},
    \label{eq:ds}
\end{equation}
where $X$ represents the $\chi^2$ test statistic with degree of freedom $k$. 
Finally, Table \ref{tab:dissimiarity_toy_example}(c) shows the dissimilarity values developed in this paper. The dissimilarity values between Trees 1 and 2, as well as between Trees 3 and 4, are relatively small, whereas the values for other pairs are notably larger.

\begin{table}
    \hspace{3.5cm}
    \begin{subtable}{0.5\textwidth}
        \centering
         \begin{tabular}{cccccc}
            \toprule
             & $X_1$ & $X_2$ & $X_3$ & $X_4$ & $X_5$ \\
             \midrule
             $I_1$ & 3 & 1 & 2 & 0 & 0 \\
             $I_2$ & 3 & 0 & 2 & 1 & 0 \\
             $I_3$ & 2 & 1 & 0 & 4 & 3 \\
             $I_4$ & 1 & 2 & 0 & 3 & 2 \\
             \bottomrule
        \end{tabular}
        \caption{Dominance}
    \end{subtable}
    
    \vspace*{0.4cm}
    \begin{subtable}{0.5\textwidth}
        \centering
         \begin{tabular}{ccccc}
            \toprule
             $X^2$ & $h_1$ & $h_2$ & $h_3$ & $h_4$ \\
             \midrule
             $h_1$ & - & 2.000\footnotesize{(3)} & 8.747\footnotesize{(4)}  & 8.215\footnotesize{(4)}   \\
             $h_2$ &  & - & 7.467\footnotesize{(4)}  & 7.875\footnotesize{(4)}   \\
             $h_3$ &  &  & - & 0.797\footnotesize{(4)}  \\
             $h_4$ &  &  &  & - \\
             \bottomrule
        \end{tabular}
        \caption{$\chi^{2}(\text{df})$ }
    \end{subtable}
    \hspace{0.4cm}
    \begin{subtable}{0.5\textwidth}
        \centering
         \begin{tabular}{ccccc}
            \toprule
             $DS$ & $h_1$ & $h_2$ & $h_3$ & $h_4$ \\
             \midrule
             $h_1$ & - & -0.192 & 1.500 & 1.386\\
             $h_2$ &  & - & 1.217 & 1.310 \\
             $h_3$ &  &  & - & -1.040 \\
             $h_4$ &  &  &  & - \\
             \bottomrule
        \end{tabular}
        \caption{Dissimilarity}
    \end{subtable}
    \caption{Procedure to calculate dissimilarity between trees}
    \label{tab:dissimiarity_toy_example}
\end{table}

\section{Simulation Study}\label{sec:Simulation}

The objective of this simulation study is to investigate three key issues. 

First, we aim to assess the extent to which HRF has a feature selection bias, as it has been reported that the bagging and RF method exhibits feature selection bias problem, due to the greedy search property of decision trees \cite{Loh2009}. 
Secondly, we seek to confirm whether HRF truly enhances diversity in comparison to bagging and RF.
Thirdly, as HRF assigns feature weights based on feature selection information from preceding trees, there is a potential risk of assigning higher selection weights to noise features. We intend to assess the severity of this problem.

Taking these factors into account, two simulation data were generated. The first simulation generated a dataset containing many features with varying numbers of unique values. The second simulation produced a dataset that included noise features.

The simulation data pertains to binary classification, comprising a total of 1,000 samples. Of these, 70\% were employed as the training data set, while the remaining 30\% constituted the evaluation data set. The ensemble consisted of 100 individual trees, and the hyper-parameters for tree creation were set to their default values. The $\alpha$ and $\beta$ parameters of HRF were set to 0.5 and 1, respectively. To ensure comprehensive comparison, the aforementioned process was repeated 100 times.

\subsection{Selection bias}
\label{sec:bias}
\begin{figure}
    \centering
    \includegraphics[width=0.75\linewidth]{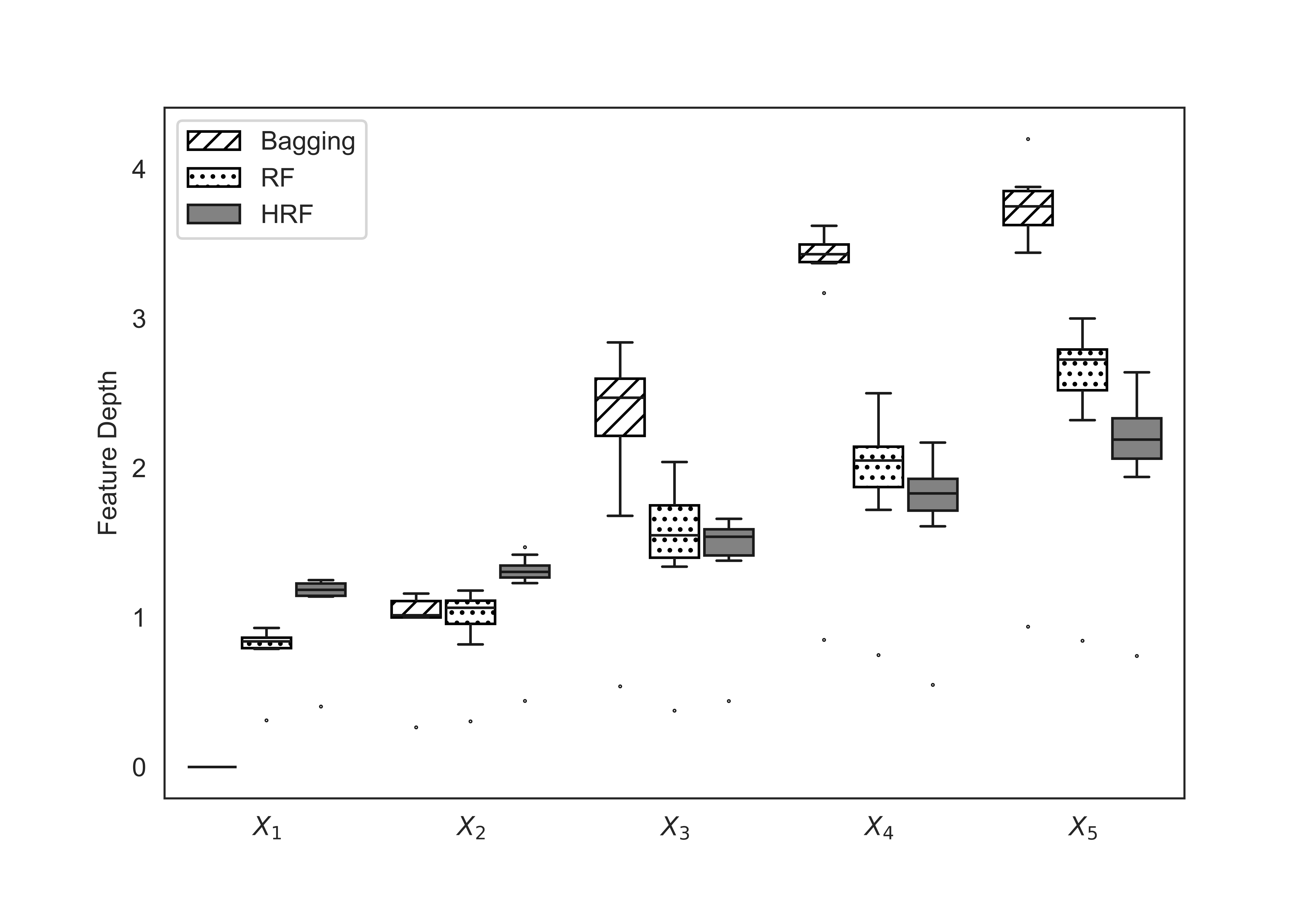}
    \caption{Box plot of feature depths}
    \label{fig:simulatoin1}
\end{figure}

In this experiment, five input features were generated. The $X_1$ feature takes integer values between 0 and 128, $X_2$ between 0 and 64, $X_3$ between 0 and 32, $X_4$ between 0 and 16, and $X_5$ between 0 and 8, all following discrete uniform distributions. The binary target variable is derived as $\boldsymbol{y} = bernoulli(\sigma(\boldsymbol{\eta} - Q_2(\boldsymbol{\eta})))$, where $Q_2(\boldsymbol{\eta})$ represents the median of $\boldsymbol{\eta}$, $\sigma(\cdot)$ denotes the sigmoid function, and $bernoulli(\cdot)$ represent the binary random variable with the given probability. The variable $\boldsymbol{\eta}$ is defined as $\boldsymbol{\eta} = X_1+X_2+X_3+X_4+X_5$, indicating that the five features are of equal importance in the classification of $\boldsymbol{y}$.

The feature depth $\boldsymbol{d}_{b} = [d_{b,1}^1,\cdots,d_{b,p}^1]$ denotes the feature depth vector at $b$-th tree in an ensemble. If there is a selection bias for features with a large number of unique values, the feature depth of $X_1$, i.e. $d_{b,1}^1$, will be smallest as it is likely to be selected earlier in the tree. Conversely, features with fewer unique values will be larger. In the event that the ensemble consists of a total of 100 trees, then 100 feature depth values will be accumulated.

Figure~\ref{fig:simulatoin1} shows box-plots representing the feature depth values for each feature across different ensemble methods.
It can be observed that bagging and RF tends to select features with a high number of unique values in the vicinity of the root node, a phenomenon that can be attributed to selection bias. Especially in the case of bagging, a significant bias issue was observed, as it almost always prioritized the selection of the $X_1$ feature.
In contrast, this bias has been considerably reduced in the case of HRF, as the feature depth values for HRF became more similar across features.

\subsection{Dissimilarity}

\begin{figure}
    \centering
    \includegraphics[width=0.75\linewidth]{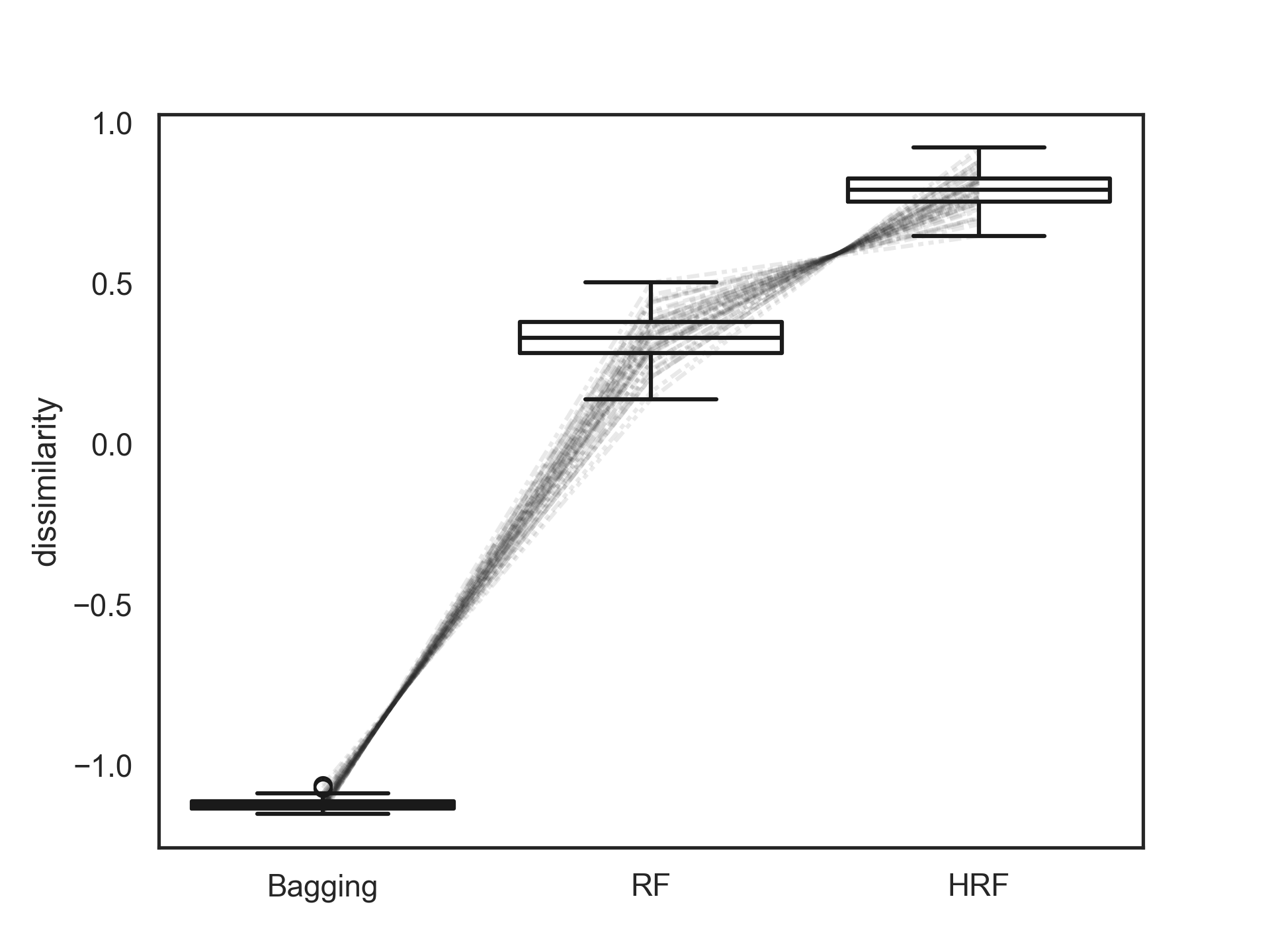}
    \caption{Dissimilarity of trees by ensemble methods.}
    \label{fig:simul1_diversity}
\end{figure}

We will discuss the results of tree dissimilarity  using the same data employed in Section~\ref{sec:bias}. 
In this experiment, we calculated the dissimilarity for the trees within the ensemble in a pairwise manner using equation~(\ref{eq:ds}). The final dissimilarity was derived as the average of all possible pairwise dissimilarity values.
This process was repeated 100 times, and the resulting line graph is presented.
As shown in Figure~\ref{fig:simul1_diversity}, we found that, as expected, the dissimilarity of bagging was the lowest, followed by RF and HRF, indicating that the diversity of HRF has improved the most.

\subsection{Noise features}
\label{sec:noise}
\begin{figure}
    \centering
    \includegraphics[width=\linewidth]{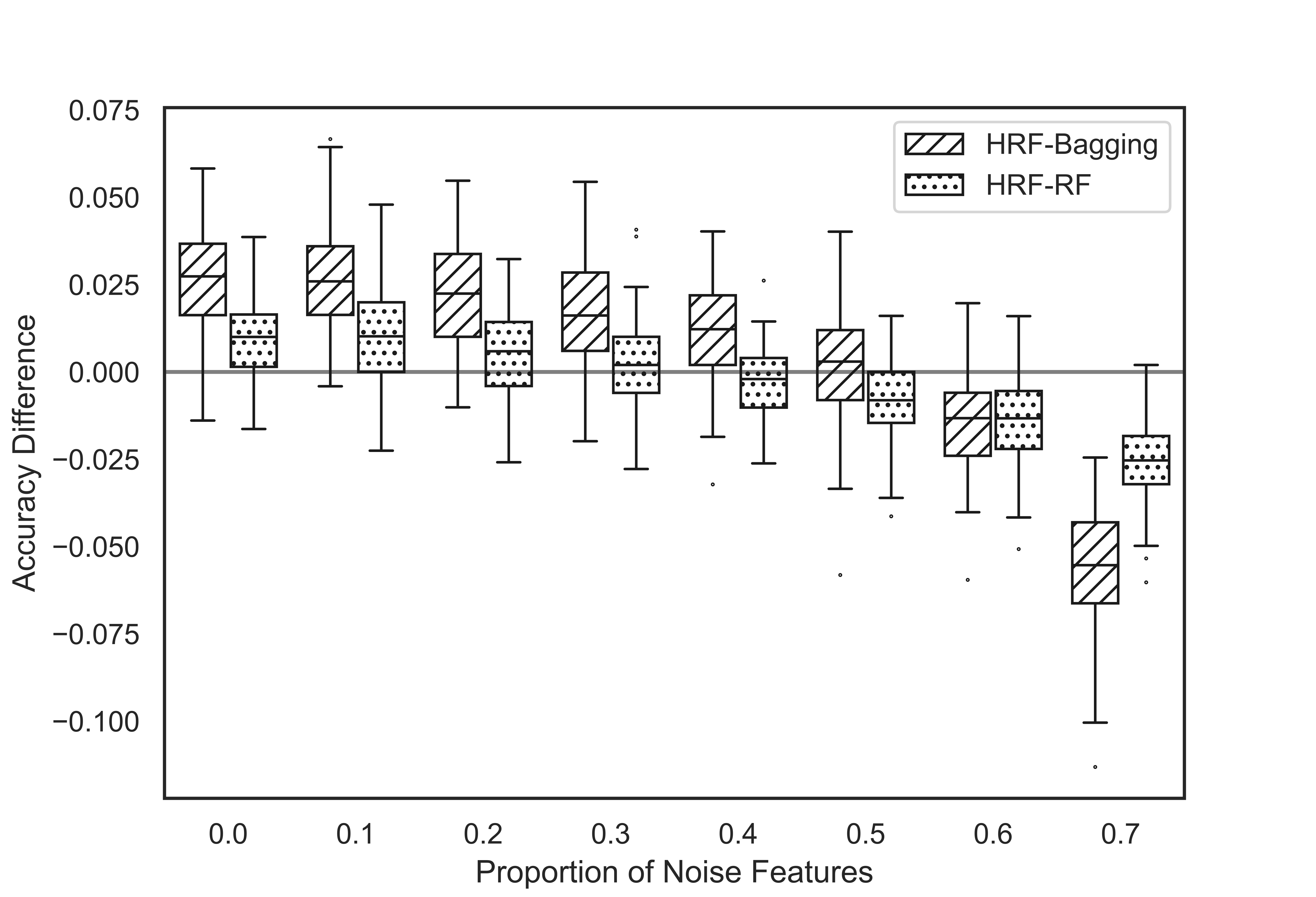}
    \caption{Accuracy differences between HRF and bagging, and between HRF and RF, based on the proportion of noise features. A positive value indicates that the HRF model exhibits superior accuracy. }
    \label{fig:simulation2}
\end{figure}

To investigate the impact of noise features on the HRF method, a case with a total of 30 features was considered. 
Let \texttt{n\char`_informative} denotes the number of informative features. The input features $x_j, j=1,\ldots,30,$ were generated from discrete uniform distribution between 0 and 4, with the size of 1,000. The target variable is generated by 
$\boldsymbol{y} = bernoulli(\sigma(\boldsymbol{\eta} - Q_2(\boldsymbol{\eta})))$, where $\boldsymbol{\eta} = \sum_{j=1}^{\texttt{n\char`_informative}} x_j$. 
In the simulation data, the noise feature ratio was increased in increments of 0.1, starting from 0 and reaching 0.9. 

Figure \ref{fig:simulation2}  presents box plots showing the accuracy differences between HRF and bagging, as well as between HRF and RF.
A positive value indicates that HRF exhibits higher accuracy.
HRF outperforms bagging when the proportion of noise features is less than 60\% of the total features. Conversely, when the proportion of noise features exceeds 60\%, HRF experiences a decline in performance. Similarly, HRF outperforms RF when the proportion of noise features is below 40\% of the total features; however, when the proportion exceeds 50\%, HRF also sees a decrease in performance.

\section{Empirical Evaluation}
\label{sec:empirical_evaluation}

In order to investigate the performance of HRF, a series of experiments were conducted. The experiments were based on 52 real or artificial datasets used in other studies or obtained from the UCI data repository (UCI) \cite{uci}. 
Table \ref{tab:data_description_clf} provides a concise overview of the data. Any missing values were filled using the mean value for numerical features and the mode value for categorical features. As an encoding method for categorical inputs, \texttt{TargetEncoder} was employed for binary classification, while \texttt{PolynomialWrapper} was utilized for multi-class classification to convert them into numerical ones \cite{McGinnis2018}.

\begin{longtable}[h]{lrrrcl}
\caption{Description of data sets}\label{tab:data_description_clf}\\ 
\toprule
 Data set & Size & Inputs & Classes & \# missing & Source \\
 \midrule
 \endfirsthead

 \multicolumn{6}{c}{\textbf{Table \ref*{tab:data_description_clf}} (Continued)}\\
 \toprule
 Data set & Size & Inputs & Classes & \# missing & Source \\
 \midrule
 \endhead

 \bottomrule
 \endfoot

 \bottomrule
 \multicolumn{6}{l}{\footnotesize $^*$ Categorical features have been encoded}\\
 \endlastfoot

 aba\footnotesize{$^*$}& 4177 & 8 & 2 & & UCI(Abalone)\\
 ail & 13750 & 12 & 2& & \cite{Loh2009} \\
 aus\footnotesize{$^*$} & 690 & 14 & 2 & &UCI(Australian credit approval)\\
 bal & 625 & 5 & 3 & &UCI(Balance scale)\\
 ban & 1672& 5 & 2 & &UCI(Bank note authentication)\\
 bcw & 683 & 9 & 2 & &\cite{lim2000comparison}\\
 bld & 346 & 6 & 2 & &UCI(BUPA liver disorders)\\
 blo & 748 & 5 & 2 & &UCI(Blood transfusion center)\\
 bod & 507 & 24 & 2 & &\cite{Heinz2003}\\
 bos\footnotesize{$^*$} & 506 & 13 & 3 & &UCI(Boston housing)\\
 bre\footnotesize{$^*$} & 699 & 10 & 2 & 16 &UCI(Wisconsin Breast Cancer)\\
 cir & 1000 & 10 & 2 & &R library mlbench(Circle in as square)\cite{leisch2010machine}\\ %
 cmc\footnotesize{$^*$} & 1473 & 9 & 3 & &UCI(Contraceptive method choice)\\
 cre\footnotesize{$^*$} & 690 & 15 & 2 & 67 &UCI(Credit approval)\\
 der & 358 & 34 & 6 & &UCI(Dermatology)\\
 dia & 768 & 8 & 2 & &\cite{Loh2009}\\
 dna\footnotesize{$^*$} & 3187 & 60 & 3 & &R package mlbench(StatLog DNA)\cite{leisch2010machine}\\ 
 eco & 336 & 7 & 4 & &\cite{Loh2009}\\
 ger\footnotesize{$^*$} & 1000 & 20 & 2 & &UCI(German credit)\\
 gla & 214 & 9 & 6 & &UCI(Glass)\\
 hea\footnotesize{$^*$} & 271 & 13 & 2 & &UCI(StatLog heart disease)\\
 hil & 606 & 100 & 2 & &UCI(Hil-valley)\\
 imp\footnotesize{$^*$} & 205 & 25 & 6 & 59 &UCI(Auto imports)\\
 int & 1000 & 10 & 2 & &\cite{KIM2011437}\\
 ion & 351 & 34 & 2 & &UCI(Ionosphere)\\
 iri & 150 & 4 & 3 & &UCI(Iris)\\
 led & 6000 & 7 & 10 & &UCI(LED display)\\
 lib & 359 & 90 & 15 & &UCI(Libra Movement)\\
 lit & 2329 & 69 & 9 & &\cite{litdata}\\
 mam & 961 & 5 & 2 & 162 &UCI(Mammographic mass)\\
 mar & 8777 & 4 & 10 & &\cite{Loh2009}\\
 mus & 8124 & 22 & 2 & &UCI(Mushroom)\\
 pid & 532 & 7 &  2 & &UCI(PIMA Indian diabetes)\\
 pks & 195 & 22 & 2 & &UCI(Parkinsons) \\
 pov & 97 & 6 & 6 & 6 &\cite{KimandLoh2001} \\
 rng & 1000 & 10 & 2 & &R library mlbench(Ringnorm)\cite{leisch2010machine}\\ %
 sat & 6435 & 36 & 6 & &UCI(StatLog satellite image)\\
 seg & 2310 & 19 & 7 & &UCI(Image segmentation)\\
 smo\footnotesize{$^*$} & 2855 & 8 & 3 & &UCI(Attitude towards smoking restrictions)\\
 snr & 208 & 60 & 2 & &R library mlbench(Sonar)\cite{leisch2010machine}\\ %
 spa & 4601 & 57 & 2 & &UCI(Spambase)\\
 spe & 267 & 44 &  2 & &UCI(SPECETF heart)\\
 tae\footnotesize{$^*$} & 151 & 5 & 3 & &KEEL(Teaching Assistant Evaluation)\cite{KEEL}\\%
 thy & 1000 & 10 & 2 & &UCI(Thyroid disease)\\
 trn & 1000 & 10 & 2 & &R library mlbench(Three norm)\cite{leisch2010machine}\\%
 twn & 1000 & 10 & 2 & &R library mlbench(Two norm)\cite{leisch2010machine}\\%
 veh & 846 & 18 & 4 & &UCI(StatLog vehicle silhouette)\\
 vol\footnotesize{$^*$} & 1512 & 6 & 6 & 40 &\cite{Loh2009}\\
 vot\footnotesize{$^*$} & 435 & 16 & 2 & &UCI(Congressional voting records)\\
 vow & 990 & 10 & 11 & &UCI(Vowel recognition)\\
 wav & 3600 & 21 & 3 & &UCI(Wave)\\
 yea & 1484 & 8 & 10 & &KEEL(Yeast)\cite{KEEL}\\%
 zoo & 101 & 16 & 7 & &R library mlbench(zoo)\cite{leisch2010machine}\\%

\end{longtable}

\subsection{Classification accuracy}     

We compared the proposed method to several ensemble methods. The candidate hyper-parameters and their sources are listed up in Table~\ref{tab:methods}. The first 3 methods are based on bagging, while the remaining ones are based on boosting. To assess the classification performance, we evaluated the classification accuracy using the results of ensemble voting. 

The experimental design is as follows: the data set was randomly split into an 80\% training set for fitting and a 20\% test set for evaluation. All methods were executed with 100 trees, with each method utilizing its optimal hyper-parameters, as determined through a 5-fold cross-validation process. HRF used the default values associated with RF for tree growth and identified the optimal $\alpha$ and $\beta$ through a 5-fold cross-validation process. In this process, $\alpha$ ranged from 0.0 to 0.9, with an increment of 0.1, while $\beta$ ranged from 1 to the minimum of 10 and $p$ (the number of features), with an increment of 1. To ensure statistical significance, the experiments were repeated 50 times and then results are assessed using the Wilcoxon signed-rank test for statistical comparison \cite{wilcoxon}.

\begin{table}[h]
    \caption{The ensemble methods and their associated hyper-parameters. The default values are in bold.}\label{tab:methods}
    \centering
    \begin{tabular}{ccc}
    \toprule
        Methods & Hyper-parameter tuning & Source \\
        \midrule
        Bagging  & \texttt{max\char`_features} = [0.6, ...,\textbf{1.0}] & \cite{Breiman1996} \cite{scikit-learn}\\ \midrule
        RF       & \texttt{max\char`_features} = [\textbf{sqrt}, log2, None] & \cite{Breiman2001}\cite{scikit-learn} \\ \midrule
        \multirow{2}{*}{ExtTree}  
                 & \texttt{max\char`_depth} = [4, 6, \textbf{{None}}] & \cite{Geurts2006}\cite{scikit-learn}\\
                 & \texttt{min\char`_samples\char`_split} = [\textbf{2}, 4] & \\ \midrule
        \multirow{3}{*}{GB}
                 & \texttt{subsample} = [0.9, \textbf{1.0}] & \cite{GB}\cite{scikit-learn} \\
                 & \texttt{max\char`_depth} = [2, \textbf{3}, 4] &        \\
                 & \texttt{min\char`_samples\char`_split} = [\textbf{2}, 4] &        \\ \midrule
        \multirow{2}{*}{XGB}
                 & \texttt{max\char`_depth} = [4, \textbf{6}, 8] & \cite{xgb}\footnotesize{$^1$}\\
                 & \texttt{min\char`_child\char`_weight} = [\textbf{1}, 2] &        \\ \midrule
        CatBoost & \texttt{default} & \cite{catboost}\footnotesize{$^2$} \\
    \bottomrule
    \end{tabular}    
    \footnotetext[1]{https://xgboost.readthedocs.io/en/stable/}
    \footnotetext[2]{https://catboost.ai/en/docs/ }
\end{table}

Table~\ref{tab:empirical_accuracy} presents a comparison of the accuracy between HRF and other ensemble methods. The '$+$' symbol indicates that HRF showed statistically more accurate results than the corresponding method, whereas the '$-$' symbol indicates less accurate results. The 'W/T/L' at the bottom row represents the number of datasets where HRF was better/equal/worse compared to the respective method.

\begin{footnotesize}
\begin{longtable}{ clllllll}
\caption{
Accuracy of ensemble methods: A '+' indicates that HRF is significantly better, while a '$-$' indicates that HRF is significantly worse at the 0.05 significance level.}
\label{tab:empirical_accuracy}\\

 \toprule
 
 {Data} & {Bagging} & {RF} & {ExtTrees} & {GB} & {XGB} & {CatBoost} & {HRF} \\ 
 \midrule
 \endfirsthead

 \multicolumn{8}{l}{\textbf{Table \ref*{tab:empirical_accuracy}}: (Continued)}\\
 \toprule
 {Data} & {Bagging} & {RF} & {ExtTrees} & {GB} & {XGB} & {CatBoost} & {HRF} \\
 \midrule
 \endhead

 \bottomrule
 \endfoot

 \bottomrule
 \multicolumn{8}{c}{\footnotesize W: \# of dataset HRF wins, T: \# of dataset tied, L: \# of dataset HRF loses}\\
 \endlastfoot
  
aba & 0.7743+ & 0.7777+ & 0.7790+ & 0.7774+ & 0.7711+ & 0.7814 & 0.7818 \\
ail & 0.8805 & 0.8803$-$ & 0.8767+ & 0.8846$-$ & 0.8838$-$ & 0.8848$-$ & 0.8798 \\
aus & 0.8692+ & 0.8676+ & 0.8647+ & 0.8661+ & 0.8698+ & 0.8671+ & 0.8765 \\
bal & 0.8706 & 0.8467+ & 0.8810$-$ & 0.9074$-$ & 0.8868$-$ & 0.8836$-$ & 0.8662 \\
ban & 0.9918+ & 0.9918+ & 0.9988$-$ & 0.9938 & 0.9945 & 0.9942 & 0.9940 \\
bcw & 0.9717+ & 0.9704+ & 0.9714+ & 0.966+ & 0.965+ & 0.971+ & 0.9746 \\
bld & 0.7089+ & 0.7175+ & 0.7146+ & 0.7258+ & 0.7045+ & 0.7159+ & 0.7375 \\
blo & 0.7657 & 0.7450+ & 0.7713 & 0.7780$-$ & 0.7551+ & 0.7918$-$ & 0.7696 \\
bod & 0.9513+ & 0.9566+ & 0.9687$-$ & 0.9617 & 0.9636 & 0.9711$-$ & 0.9621 \\
bos & 0.7911+ & 0.7921+ & 0.7796+ & 0.7887+ & 0.7853+ & 0.7887+ & 0.8005 \\
bre & 0.9656+ & 0.9670+ & 0.9685 & 0.9615+ & 0.9581+ & 0.9684+ & 0.9701 \\
cir & 0.8263+ & 0.8269+ & 0.8291 & 0.8699$-$ & 0.8781$-$ & 0.8618$-$ & 0.8340 \\
cmc & 0.5191+ & 0.5167+ & 0.5227+ & 0.5566$-$ & 0.5372$-$ & 0.5318 & 0.5300 \\
cre & 0.8731+ & 0.8742+ & 0.8656+ & 0.8671+ & 0.8678+ & 0.8746+ & 0.8825 \\
der & 0.9677+ & 0.9744+ & 0.9817 & 0.9637+ & 0.9602+ & 0.9753+ & 0.9798 \\
dia & 0.7562+ & 0.7624+ & 0.7647+ & 0.7616+ & 0.7430+ & 0.7718 & 0.7706 \\
dna & 0.9621+ & 0.9678 & 0.9670 & 0.9659+ & 0.9663+ & 0.9656+ & 0.9679 \\
eco & 0.9258+ & 0.9368+ & 0.9442 & 0.9106+ & 0.9154+ & 0.9308+ & 0.9450 \\
ger & 0.7648+ & 0.7678+ & 0.7648+ & 0.7665+ & 0.7589+ & 0.7745 & 0.7728 \\
gla & 0.7438+ & 0.7581+ & 0.7403+ & 0.7347+ & 0.7319+ & 0.7541+ & 0.7688 \\
hea & 0.8195+ & 0.8286+ & 0.8323+ & 0.8062+ & 0.8079+ & 0.8328+ & 0.8407 \\
hil & 0.5674+ & 0.5625+ & 0.5536+ & 0.5579+ & 0.5744 & 0.5118+ & 0.5781 \\
imp & 0.7885 & 0.7810+ & 0.7351+ & 0.7915 & 0.7987 & 0.7849 & 0.7895 \\
int & 0.7455$-$ & 0.5946 & 0.5571+ & 0.6901$-$ & 0.8469$-$ & 0.8050$-$ & 0.5994 \\
ion & 0.9310+ & 0.9345 & 0.9394 & 0.9295+ & 0.9274+ & 0.9328 & 0.9375 \\
iri & 0.9449+ & 0.9493+ & 0.9551 & 0.9436+ & 0.9378+ & 0.9489+ & 0.9511 \\
led & 0.7340+ & 0.7315+ & 0.7334+ & 0.7346+ & 0.7323+ & 0.7311+ & 0.7354 \\
lib & 0.7748+ & 0.7811+ & 0.8141$-$ & 0.6678+ & 0.7119+ & 0.7809+ & 0.8006 \\
lit & 0.8808+ & 0.8835+ & 0.8815+ & 0.8898 & 0.9055$-$ & 0.9033$-$ & 0.8878 \\
mam & 0.8295$-$ & 0.7938+ & 0.8227 & 0.8247 & 0.8081+ & 0.8350$-$ & 0.8217 \\
mar & 0.5812 & 0.5400+ & 0.5815$-$ & 0.5803+ & 0.5751+ & 0.5678+ & 0.5812 \\
pid & 0.7596+ & 0.7706+ & 0.7618+ & 0.7711 & 0.7390+ & 0.7772 & 0.7767 \\
pks & 0.8738+ & 0.8969+ & 0.8679+ & 0.9059 & 0.9128 & 0.9166$-$ & 0.9055 \\
pov & 0.6090+ & 0.6117 & 0.5869+ & 0.6048 & 0.6228 & 0.6386 & 0.6228 \\
rng & 0.9179 & 0.9085+ & 0.9374$-$ & 0.9161 & 0.9169 & 0.9231$-$ & 0.9161 \\
sat & 0.9116+ & 0.9139+ & 0.9144 & 0.9088+ & 0.9153 & 0.9044+ & 0.9151 \\
seg & 0.9777 & 0.9778 & 0.9775 & 0.9792 & 0.9822$-$ & 0.9777 & 0.9783 \\
smo & 0.6986$-$ & 0.6511+ & 0.6983$-$ & 0.6916$-$ & 0.662+ & 0.6480+ & 0.6835 \\
snr & 0.7910+ & 0.8216+ & 0.8265+ & 0.8329+ & 0.8432 & 0.8416 & 0.8439 \\
spa & 0.9499+ & 0.9518+ & 0.9548$-$ & 0.9432+ & 0.9537$-$ & 0.9528 & 0.9527 \\
spe & 0.7990+ & 0.8032+ & 0.8040+ & 0.7825+ & 0.7953+ & 0.808+ & 0.8155 \\
tae & 0.6138+ & 0.6333+ & 0.6093+ & 0.6396 & 0.6573$-$ & 0.6342 & 0.6440 \\
thy & 0.9965$-$ & 0.9963$-$ & 0.9811+ & 0.9961 & 0.9953+ & 0.9960 & 0.9961 \\
trn & 0.8538+ & 0.8596+ & 0.8736$-$ & 0.8551+ & 0.8556+ & 0.8604+ & 0.8659 \\
twn & 0.9627+ & 0.9623+ & 0.9672$-$ & 0.9613+ & 0.9587+ & 0.9643 & 0.9657 \\
veh & 0.7477+ & 0.7527+ & 0.7452+ & 0.7636 & 0.7645 & 0.7645 & 0.7589 \\
vol & 0.8625+ & 0.9036 & 0.8893+ & 0.9021 & 0.9041 & 0.9073$-$ & 0.9035 \\
vot & 0.9548+ & 0.9548+ & 0.9551+ & 0.9612 & 0.9568 & 0.9517+ & 0.9594 \\
vow & 0.9379+ & 0.9467+ & 0.9733$-$ & 0.8782+ & 0.8936+ & 0.9421+ & 0.9551 \\
wav & 0.8465+ & 0.8466+ & 0.8541 & 0.8488+ & 0.8474+ & 0.8462+ & 0.8516 \\
yea & 0.5933+ & 0.6098+ & 0.6062+ & 0.5953+ & 0.5808+ & 0.5873+ & 0.6201 \\
zoo & 0.9473 & 0.9447 & 0.9393+ & 0.9433 & 0.9447 & 0.9427 & 0.9507 \\ \midrule
W/T/L & (40/8/4) & (43/7/2) & (29/12/11) & (29/16/7) & (30/13/9) & (24/17/11)& ~\\

\end{longtable}
\end{footnotesize}

Table \ref{tab:comparison_table} displays pairwise comparisons among the methods. The values in the table show the number of instances where the method listed vertically is more accurate than the method listed horizontally, with the number in parentheses denoting statistical significance. To illustrate, a value of 35(22) in the first row and second column indicates that RF outperforms the Bagging method in 35 out of 52 datasets, with 22 of those instances being statistically significant. In cases where the accuracy values are the same, a value of 0.5 was assigned.

\begin{table}[h]
\caption{Summary of comparisons among methods}
\label{tab:comparison_table}
    \centering
    \begin{tabular}{cccccccc}
    \toprule
    ~ & {Bagging} & {RF} & {ExtTree} & {GB} & {XGB} & {CatBoost} & {HRF} \\
    \midrule
    {Bagging}  &        & 35(22) & 32.5(25) & 26(21) & 24(19) & 37(30) & 45(40) \\
    {RF}       & 17(8)  &        & 30(19) & 24(17) & 28.5(23) & 32(22) & 48(43) \\
    {ExtTree}  & 19.5(12) & 22(14) &        & 27(18) & 22(15) & 30(21) & 35(29) \\
    {GB}       & 26(16) & 28(16) & 25(21) &        & 27(14) & 33.5(28) & 37(29) \\
    {XGB}      & 28(22) & 23.5(21) & 30(24) & 25(20) &        & 30.5(26) & 34.5(30) \\
    {CatBoost} & 15(7)  & 20(8)  & 22(19) & 18.5(11) & 21.5(11) &        & 33(24) \\
    {HRF} & 7(4)   & 4(2)   & 17(11) & 15(7)  & 17.5(9)  & 19(11) & \\
    \bottomrule
    \end{tabular}
    \footnotetext{In each cell, the results are summarized as '$a(b)$', where $a$ indicates the number of datasets where the method in the column outperforms the method in the row, and $b$ indicates the number of datasets where this difference is statistically significant according to the one-sided paired $t$-test.}
\end{table}

Table \ref{tab:dominance_rank} summarises the ranking of methods according to the results presented in Table \ref{tab:comparison_table}. 
The dominance rank is calculated as the difference between the number of significant wins and significant losses. For example, HRF achives a dominance rank of 151, derived from a total of 195 significant wins and 44 significant losses. The number of significant wins for HRF is equal to the cumulative sum of the values within parentheses in the HRF columns of Table \ref{tab:comparison_table}. Similarly, the number of significant losses for HRF is the sum of the values within parentheses in the HRF row. Tables~\ref{tab:comparison_table} and \ref{tab:dominance_rank} collectively demonstrate that the HRF method exhibits significant superiority over other methods across a dataset pool comprising 52 instances. 

\begin{table}
\caption{The dominance rank of the methods determined by the significant difference from the results in Table~\ref{tab:comparison_table}}
\label{tab:dominance_rank}
    \centering
    \begin{tabular}{cccc}
    \toprule
        {Methods} &{Wins} & {Losses} & {Dominance Rank}  \\ \midrule
        HRF & 195 & 44  & 151\\
        CatBoost & 138 & 80 & 58\\
        ExtTree  & 119 & 109 & 10\\
        GB      & 94 & 124 & -30 \\
        RF       & 83 & 132 & -49 \\
        XGB       & 91 & 143 & -52 \\
        Bagging  & 69 & 157 & -88 \\ \bottomrule
    \end{tabular}
\end{table}

\subsection{Noise feature effect}\label{subsec:ExcelCondition}
In Section~\ref{sec:noise}, we discovered the effects of noise features on HRF. Specifically, if there are too many noise features, the accuracy becomes worse than that of bagging and RF. In this section, we examined whether a similar pattern exists in experiments with real data. To estimate the number of noise features in a dataset, a deep decision tree was constructed and the feature importance for each feature was calculated. Let's denote the feature importance vector as $\boldsymbol{I} = [I_1, ..., I_p]$. We normalize $\boldsymbol{I}$ such that $\boldsymbol{I} \leftarrow [I_1, ..., I_p] / \sum I_j$. If the scaled feature importance is less than $1/p$, i.e. $I_j < \frac{1}{p}$, we designate that feature as noise. Finally, the proportion of noise features (\texttt{p\char`_noise}$=\frac{\# \text{noise features}}{p}$) is summarized in Table \ref{tab:data_column_prop}. 

\begin{figure}
    \centering
    \includegraphics[width=0.75\linewidth]{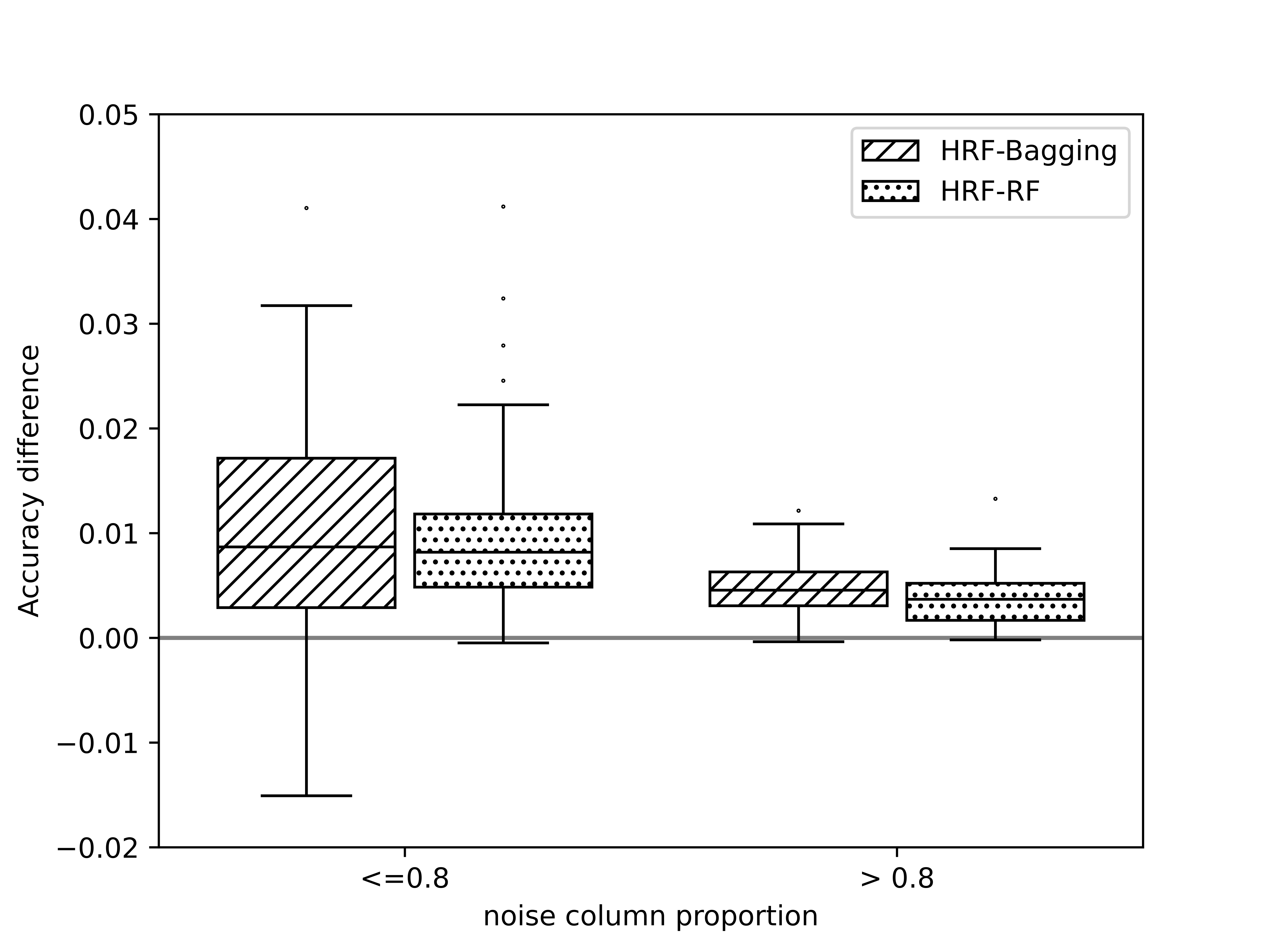}
    \caption{Box plots for the accuracy difference between HRF and others}
    \label{fig:empirical_boxplot}
\end{figure}

We divided the entire dataset into two groups based on a noise ratio of 0.8.
Then, using the accuracy differences between HRF and bagging, as well as between HRF and RF within each group, we created box plots, which are shown in Figure~\ref{fig:empirical_boxplot}.
The accuracy difference between HRF and other ensemble methods was greater in the group with a lower proportion of noise features.

\begin{table}
\caption{The estimated proportions of noise features}\label{tab:data_column_prop}
    \centering
    \begin{tabular}{cc|cc|cc}
    \hline
    \textbf{Data set} & \texttt{p\char`_noise} & \textbf{Data set} & \texttt{p\char`_noise} & \textbf{Data set} & \texttt{p\char`_noise} \\ \hline
    aba & 0.63 & ger & 0.65 & sat & 0.92  \\ 
    ail & 0.75 & gla & 0.44 & seg & 0.79  \\ 
    aus & 0.79 & hea & 0.69 & smo & 0.7  \\ 
    bal & 0.25 & hil & 0.6 & snr & 0.75  \\ 
    ban & 0.75 & imp & 0.78 & spa & 0.79  \\ 
    bcw & 0.89 & int & 0.8 & spe & 0.7  \\ 
    bld & 0.67 & ion & 0.85 & tae & 0.71  \\ 
    blo & 0.5 & iri & 0.75 & thy & 0.81  \\ 
    bod & 0.79 & led & 0.43 & trn & 0.4  \\ 
    bos & 0.71 & lib & 0.68 & twn & 0.6  \\ 
    bre & 0.89 & lit & 0.77 & veh & 0.67  \\ 
    cir & 0.7 & mam & 0.6 & vol & 0.71  \\ 
    cmc & 0.85 & mar & 0.75 & vot & 0.94  \\ 
    cre & 0.73 & pid & 0.57 & vow & 0.7  \\ 
    der & 0.82 & pks & 0.73 & wav & 0.71  \\ 
    dia & 0.75 & pov & 0.67 & yea & 0.5  \\ 
    dna & 0.9 & rng & 0.5 & zoo & 0.69  \\ 
    eco & 0.71 & ~ & ~ & ~ & ~ \\ \hline 
    \end{tabular}
\end{table}

\section{Conclusion}
\label{sec:conlusion}

In this paper, we proposed an ensemble method designed to modify RF to enhance diversity in decision tree structures. Our method demonstrated superior performance compared to other ensemble methods across most benchmark datasets.
The idea is to reduce the probability that key features used in the previous trees will be selected again in the next tree.
Unlike RF, which tends to favor features with many unique values, we found that HRF also helps mitigate this selection bias.
However, excessive noise features can have a negative impact on HRF performance. As demonstrated by the simulations and empirical results, HRF performs well on datasets with low noise and a diverse range of unique feature values.

Additionally, this paper introduces a methodology to quantify diversity within an ensemble by analyzing feature dominance, allowing for a comparison of the diversity across ensemble methods. Overall, it was found that the HRF ensemble method comprised more diverse trees than traditional methods, leading to more accurate classification predictions.

\section*{Declarations}

\noindent\textbf{Funding} 
Hyunjoong Kim’s work was supported by the IITP(Institute of Information \& Coummunications Technology Planning \& Evaluation)-ICAN(ICT Challenge and Advanced Network of HRD) grant funded by the Korea government(Ministry of Science and ICT)(IITP-2023-00259934) and by the National Research Foundation of Korea (NRF) grant funded by the Korean government (No. 2016R1D1A1B02011696).

\end{document}